# Un modèle générique d'organisation de corpus en ligne : application à la *FReeBank*

S. Salmon-Alt* — L. Romary** — J.-M. Pierrel*

*ATILF – UMR 7118*
*Analyse et Traitement Informatique de la Langue Française*
*44, avenue de la Libération, BP 30687, F-54063 Nancy Cedex*
*{salt@atilf.fr, Jean-Marie.Pierrel@atilf.fr}*

**\*\*** *LORIA – UMR 7503*
*Laboratoire Lorrain de Recherche en Informatique et ses Applications*
*Campus scientifique, BP 239, F-54506 Vandœuvre-lès-Nancy Cedex*
*Laurent.Romary@loria.fr*

RÉSUMÉ. *Les corpus français librement accessibles et annotés linguistiquement sont insuffisants à la fois quantitativement et qualitativement. Partant de ce constat, la FREEBANK se veut une base de corpus du français annotés à plusieurs niveaux (structurel, morphologique, syntaxique, coréférentiel) et à différents degrés de finesse linguistique qui soit libre d'accès, codée selon des schémas normalisés, intégrant des ressources existantes et ouverte à l'enrichissement progressif. Préalablement à la présentation du prototype qui a été réalisé, le présent article propose une modélisation générique de l'organisation et du déploiement d'une archive de corpus linguistiques dans la continuité des travaux menés au niveau international sur la représentation des ressources linguistiques (TEI et ISO/TC 37/SC 4).*

ABSTRACT. *The few available French resources for evaluating linguistic models or algorithms on other linguistic levels than morpho-syntax are either insufficient from quantitative as well as qualitative point of view or not freely accessible. Based on this fact, the FREEBANK project intends to create French corpora constructed using manually revised output from a hybrid Constraint Grammar parser and annotated on several linguistic levels (structure, morpho-syntax, syntax, coreference), with the objective to make them available on-line for research purposes. Therefore, we will focus on using standard annotation schemes, integration of existing resources and maintenance allowing for continuous enrichment of the annotations. Prior to the actual presentation of the prototype that has been implemented, this paper describes a generic model for the organization and deployment of a linguistic resource archive, in compliance with the various works currently conducted within international standardization initiatives (TEI and ISO/TC 37/SC 4).*

MOTS-CLÉS : *ressources linguistiques, annotation multi-niveau, normalisation, ressources libres…*

KEYWORDS: *linguistic resources, multi-level annotation, standardization, open resources…*

**1. Introduction**

L'idée de la *FReeBank* – en tant qu'espace de dépôt, de maintenance, de distribution et de normalisation de ressources libres pour l'étude et le traitement du Français – fait partie des résultats théoriques et pratiques majeurs d'un projet précédent (*Ananas*[1] ; Salmon-Alt, 2002). Ce projet, centré initialement sur l'annotation sémantique de corpus existants, s'est en effet rapidement heurté à l'indisponibilité de corpus pré-annotés, réutilisables et libres de droit. Au-delà d'un important travail de collecte et d'annotation de corpus libres (balisage TEI (Sperberg-McQueen et Burnard 2002), segmentation, étiquetage morpho-syntaxique, analyse syntaxique et annotation anaphorique), cela nous a amenés à réfléchir d'une façon plus générale à une architecture de gestion de ressources linguistiques en ligne. En plus de notre volonté d'encourager le partage de corpus annotés dans la communauté scientifique, l'initiative de la *FReeBank* a été motivée par trois constats positifs : le déploiement de plusieurs campagnes d'annotation, manuelle ou automatique, au-delà du niveau morpho-syntaxique (projets Evalda Easy/Média[2]), d'autres initiatives de mise en ligne ou de recensement de corpus libres (Asila[3], ABU[4], BDCOIFA[5]) et l'avancement des fondements théoriques de la normalisation de ressources linguistiques (Bird et Liberman, 2001 ; Ide et Romary, 2004b et les travaux de l'ISO TC 37/SC 4[6]). Dans ce contexte, la *FReeBank* se veut un espace ouvert de gestion de ressources libres, permettant le dépôt et le téléchargement de données brutes ou annotées, mais aussi le dépôt d'annotations sur des ressources existantes, par exemple sous forme de méta-annotations (validation, dépréciation, correction ou affinage d'annotations précédentes) ou d'annotations concurrentes (annotations multi-annotateur). Si elle n'impose aucune validation *a priori* des annotations soumises, elle met l'accent sur une documentation exhaustive et standardisée des données répertoriées. Cela concerne d'une part la documentation des informations linguistiques apportées par l'annotation et, d'autre part, la documentation des méta-données, en particulier la description de la ressource, des niveaux de description et des contributeurs. Dans cet esprit, la *FReeBank* est conçue comme un espace expérimental ouvert, évolutif et générique, fondé sur une méthodologie de modélisation d'annotations linguistiques qui tente d'allier une analyse fine des pratiques et besoins linguistiques aux initiatives internationales de représentation de données langagières.

---

[1] http://www.atilf.fr/ananas
[2] http://www.technolangue.net/
[3] http://www.loria.fr/projets/asila/
[4] http://abu.cnam.fr/
[5] http://www.unicaen.fr/corpus/
[6] http://www.tc37sc4.org

## 2. Fondements théoriques

### 2.1. *Organisation générique d'une archive linguistique*

#### 2.1.1. La notion de « corpus »

La *FreeBank* repose sur l'idée fondamentale que toute donnée linguistique peut se caractériser par une certaine couverture linguistique, c'est-à-dire un contenu langagier linéaire fini identifié dans un contexte de production particulier. Le contexte de production peut être spécifié de multiples façons. Il peut s'agir par exemple, pour l'écrit, d'une édition particulière d'une œuvre, ou, pour l'oral, de l'identification d'un (ou de plusieurs) locuteur(s) et d'un instant de locution. À partir de cette notion, la *FreeBank* définit un corpus comme une collection de données relatives à une certaine couverture linguistique, vue en tant qu'objet d'études linguistiques ou littéraires. Ce concept couvre, par exemple, des sources audio d'une conversation enregistrée, des textes écrits bruts, des dialogues transcrits ou encore du matériau linguistique existant sous forme de manuscrits anciens numérisés. Dans cette acception, le texte intégral du *Père Goriot* de Balzac, dans l'édition de Gallimard 1976, constitue un corpus différent de celui obtenu par la sélection du chapitre I de la même édition de cet ouvrage. À l'opposé, le texte du chapitre 1 du *Père Goriot* étiquetée morpho-syntaxiquement (Figure 5) et ce même texte annoté en anaphores (Figure 8) seront considérés comme relatifs à un même corpus. Notre notion de corpus est volontairement générale par rapport à certaines positions dans le domaine (McEnery et Wilson 1996, Habert et al. 1997, Véronis 2000). Parmi les critères proposés par McEnery et Wilson (1996), elle reprend celui de la « taille finie » et celui de la « disponibilité sous forme électronique ». En contrepartie, s'appuyant sur des arguments développés en particulier par Kilgarriff et Grefenstette (2003) à propos de la difficulté d'évaluation de la « représentativité » d'un corpus, elle couvre des compilations linguistiques intentionnelles aussi bien que contingentes.

#### 2.1.2. La difficulté de définir l' « annotation linguistique »

La valeur ajoutée d'un corpus linguistique augmente avec le nombre et la qualité des annotations. D'une façon générale, l'annotation consiste à expliciter des informations linguistiques jusqu'alors implicites dans le matériau, en y ajoutant des données méta-linguistiques (Bird et Liberman, 2001 ; McEnery et Wilson, 1996). McEnery et Wilson (1996) distinguent sept types d'annotations linguistiques : l'étiquetage morpho-syntaxique (Figures 4 et 5), la lemmatisation (Figures 4 et 5), l'annotation syntaxique (Figures 6 et 7), l'annotation sémantique (Figure 7), l'annotation discursive (Figure 8) et la transcription phonétique. Si cette classification va fortement dans le sens du postulat selon lequel une annotation linguistique est toujours le résultat d'un processus d'interprétation (Leech 1993), elle soulève aussi certains problèmes. En particulier, elle exclut certains types d'ajouts d'informations, comme la segmentation en chaînes de caractères (Figure 3), l'explicitation de la structure d'un texte à la TEI (Figure 2) ou l'insertion de bornes temporelles dans une transcription. Si l'on peut en effet considérer que ces cas ne relèvent pas de l'interprétation – notamment parce qu'il n'y a pas de possibilité de divergence entre annotateurs –  ils sont toutefois souvent considérés comme annotations, simplement parce qu'il y a explicitation d'informations préalablement implicites (Bird et Liberman, 2001). Plus généralement, comme le font remarquer à juste titre Leech (1997) et Véronis (2000), il peut être difficile de départager ce qui relève de la représentation de ce qui relève de l'interprétation. Ceci est particulièrement vrai pour les transitions entre « recueil de données » et « transcription », où la part de l'interprétation peut être importante, par exemple lors de la transcription d'enregistrement oraux ou lors du balisage de divisions hiérarchisées dans un texte. Enfin, il peut très bien exister toute une série de représentations successives avant même d'arriver à une annotation au sens classique (fichier audio, transcription phonétique, transcription orthographique, balisage structurel de base à la TEI, segmentation, annotation morpho-syntaxique). Vouloir, dans ces cas, maintenir une séparation entre « représentations » et « annotations » poserait non seulement la question du choix du « matériau primaire » (c'est-à-dire celui qui est supposé servir de référence objective vis-à-vis de la couverture linguistique et donc de toute autre activité d'annotation), mais aussi celle de la justification du statut particulier attribué à celui-ci.  Ainsi, si l'on tient vraiment à identifier ce matériau primaire, il devient difficile de choisir entre une stratégie qui sélectionnerait un ensemble hétérogène de niveaux les plus « bruts » parmi ceux disponibles (fichier audio, transcription orthographique, et pourquoi pas annotation morpho-syntaxique...), au risque d'y introduire des données subjectives, ou le choix forcé d'un niveau particulier de représentation, au risque que celui-ci ne soit pas toujours présent dans un corpus donné (par exemple lorsque l'on ne dispose que de la transcription dans le cas de données orales).

*2.1.3. De l' « annotation » aux « niveaux de description »*

Partant de ces constats – et en radicalisant en quelque sorte encore la position adoptée par Véronis (2000) ainsi que par Bird et Liberman (2001) – nous proposons d'unifier les notions de « recueil de données », « transcription » et « annotation » par l'introduction de la notion unifiée de « niveau de description ». Nous entendons par là tout ensemble cohérent d'informations explicites relatif à un corpus (au sens de notre définition). Par rapport à la notion d'annotation discutée ci-dessus, nous maintenons donc le critère d'explicitation d'information, mais nous n'imposons plus de contraintes sur la nature de ces informations (elles peuvent être linguistiques, mais aussi temporelles ou structurelles), ni sur la part de l'interprétation dans ce processus. Dans cette acception, un niveau de description, toujours relatif à un corpus donné, couvre tout aussi bien un enregistrement audio, du texte brut ou formaté (Figure 1), du texte balisé structurellement (Figure 2), du texte segmenté (Figure 3) et des « annotations » au sens classique, quelqu'en soit le format (Figure 4 à 8).  D'un point de vue plus pratique, la question de l'existence d'un « matériau brut » est reposée dans un contexte différent : plutôt que de nous interroger sur les critères informationnels permettant d'attribuer un statut privilégié à tel ou tel ensemble de données, nous nous interrogeons sur la dépendance entre niveaux de description.

Pour la *FReeBank*, nous avons ainsi identifié que l'une des informations essentielles qui devait être attachée à un niveau de description donné (ou un groupement de niveaux de description dans le cas d'annotations internes, cf. *supra*), est son lien de dépendance éventuel vis-à-vis d'un autre niveau de description. Ce lien établit que le niveau de description doit être complété par les informations d'autres niveaux (éventuellement par transitivité) pour être parfaitement exploitable. Ainsi, on pourra marquer la dépendance entre une transcription orthographique et le fichier audio qui lui a servi de source, ou encore entre une annotation référentielle basée sur une annotation syntaxique identifiant les groupes nominaux (Salmon-Alt et Romary, 2005).

Combiné avec la notion de couverture linguistique, la dépendance entre deux niveaux de description permet maintenant de fournir un cadre plus rigoureux à la définition de ce que l'on peut identifier, au sein d'un corpus donné, comme étant du matériau primaire ou secondaire, et ce par le biais d'une propriété caractérisant spécifiquement un niveau de description. Un niveau de description est ainsi considéré comme « secondaire », dès lors qu'il est nécessaire de faire référence à un autre niveau de description avec lequel il est en dépendance pour en reconstruire la couverture linguistique. Ce sera par exemple le cas d'un étiquetage morpho-syntaxique (FIgure 5) qui pointerait, sans la dupliquer, sur une transcription phonétique ou une segmentation en unité de référence (Figure 3). La reconstitution de la couverture peut se faire par transitivité, l'essentiel étant un continuum de dépendance d'un niveau de description secondaire vers, à un moment donné, un niveau primaire. Tout niveau qui n'est pas strictement dépendant d'un autre pour la reconstitution de la couverture linguistique sera alors considéré comme « primaire ». Il faut toutefois noter que cette notion dépend *in fine* de ce que l'on considère comme la couverture linguistique. Par exemple, une transcription d'un enregistrement audio, alignée par des bornes temporelles avec l'information sonore (cf. recommandations de la TEI), permet de reconstruire partiellement la couverture linguistique du corpus (d'un point de vue orthographique), tout en comportant des pointeurs sur un autre niveau de description qui complète la connaissance que l'on peut avoir de cette même couverture linguistique. Dans ce cas, les deux niveaux de description sont considérés comme étant primaires. Il peut en être de même pour une annotation référentielle dans laquelle des segments linguistiques (les expressions référentielles) sont reliés à des représentations identifiantes pour des objets du contexte (Anderson et al., 1991 ; Salmon-Alt, 2001).

Par ailleurs, on ne peut introduire les notions de niveau de description et de dépendance sans préciser les conditions de représentation effective d'un ensemble de niveaux de descriptions au sein d'un même corpus : un niveau de description donné peut ainsi être physiquement intégré ou superposé à un autre niveau de description ou représenté de façon indépendante. Dans ce dernier cas, s'il y a des liens de dépendance vers un autre niveau de description, il faudra envisager des mécanismes de référenciation (pointeurs). Ces deux modes de représentation correspondent aux deux notions d' « annotation interne » (*inline markup*) et d' « annotation externe » (*stand-off markup*), introduites au milieu des années 90 par plusieurs auteurs (Ide et Véronis, 1995; Ide et Priest-Dorman, 1996; Thomson et McKelvie, 1997). Une annotation interne est une représentation simultanée de plusieurs niveaux de description au sein d'un même objet informationnel : par exemple, structure, sémantique et coréférence (Figure 2), morpho-syntaxe et syntaxe (Figure 6) ou syntaxe et sémantique (Figure 7). Elle introduit de fait des relations d'ordre et de hiérarchisation entre éléments informationnels appartenant à différents niveaux de description. Or, ces relations peuvent être intentionnelles, mais aussi contingentes : dans un court dialogue de renseignements, il se pourrait, par exemple, que les tours de parole correspondent systématiquement à des actes dialogiques, sans que cela puisse être considéré comme une relation générale et systématique. À l'opposé, l'annotation externe sépare physiquement les niveaux de description et explicite de ce fait les relations de dépendance éventuelles entre ceux-ci par des mécanismes de référence, par exemple des pointeurs (Figures 3 et 5). Pour l'exemple dialogique précédent, l'introduction d'une dépendance explicite entre tours de parole et actes de dialogue est alors peu probable et irait à l'encontre du bon sens linguistique : cela empêcherait par exemple de tenir compte d'un acte dialogique réparti sur deux tours de parole ou d'un tour de parole associé, simultanément ou non, à plusieurs actes dialogiques. En contrepartie, introduire une dépendance entre les niveaux de description morpho-syntaxique et syntaxique semble approprié, puisque la délimitation des constituants syntaxiques repose fondamentalement sur la segmentation opérée au niveau morpho-syntaxique. En pratique, cela se traduira alors pour un pointage de l'annotation vers les unités de référence issues de l'annotation morphologique.

Parmi ces deux modes de représentation, l'annotation externe est le mode le plus générique, généralement recommandé pour une meilleure gestion d'un corpus (Mengel et al., 2000 ; Ide et Romary, 2004b). En raison de l'explicitation des dépendances entre niveaux de description, il permet en effet la représentation parallèle d'un nombre arbitraire de niveaux de description, éventuellement non hiérarchiques, autorise la co-existence de plusieurs versions concurrentes d'un même niveau de description (section 3.2.1.) et permet de modifier, voire supprimer des informations sur un niveau de description particulier sans rendre inutilisables les autres informations. Dans le contexte de la *FReeBank*, où l'intention est justement de ne pas imposer, *a priori*, un modèle particulier d'annotation à toute la communauté, nous donnons la possibilité de déposer des données reposant sur l'un ou l'autre modèle, même au sein d'un même corpus.

*2.1.4. L'unité de dépôt physique : la « ressource »*

Enfin, l'organisation conceptuelle (ou linguistique) d'un corpus en niveaux de description se superpose avec son organisation physique. D'un point de vue physique, un corpus déposé à la *FreeBank* est organisé en *ressources*. La ressource est l'unité de dépôt, c'est-à-dire le fichier électronique soumis par un dépositaire. Il convient de noter qu'il n'y a pas d'isomorphisme entre l'organisation conceptuelle et l'organisation physique d'un corpus. Un même corpus peut en effet se présenter sous forme d'une seule ressource, comportant simultanément plusieurs niveaux de description : c'est par exemple le cas de la sortie de l'analyseur syntaxique *VISL-FraG*[7] (Bick, 2003), comportant des informations strictement syntaxiques (constituants et dépendances), mais aussi les résultats d'un étiquetage morpho-syntaxique et d'une lemmatisation (Figure 6). A l'opposé, les informations relatives à un même niveau d'annotation peuvent être « éparpillées » sur plusieurs ressources : c'est par exemple le cas de la sortie du résolveur d'anaphores *Art Nouveau* (Vieira et al., 2003), séparant en unités de dépôts distinctes les listes des expressions anaphoriques, des antécédents et des liens anaphoriques.

| **Texte brut** |
|---|
| Madame Vauquer, née De Conflans , est une vieille femme qui, depuis quarante ans, tient à Paris une pension bourgeoise établie rue Neuve-Sainte-Geneviève , entre le quartier latin et le faubourg Saint-Marceau. Cette pension, connue sous le nom de la Maison-Vauquer , admet également des hommes et des femmes, des jeunes gens et des vieillards, sans que jamais la médisance ait attaqué les mœurs de ce respectable établissement. |

**Figure 1.** *« Goriot » (extrait) : texte source*

| **Structure, entités nommées et anaphores** |
|---|
| \<p>\<seg>\<rs type="person-oeuvre" id="p1">\<name type="person-oeuvre" key="Mme Vauquer">Madame Vauquer\</name>, née\<name type="person-oeuvre" key="De Conflans">De Conflans\</name>\</rs>, est une vieille femme qui, depuis quarante ans, tient à \<rs type="place-ville" id="pl1">\<name type="place-ville" key="Paris">Paris\</name>\</rs> \<rs type="org-oeuvre" id="or1">une pension bourgeoise établie\<rs type="place-rue" id="pl2">\<name type="place-rue" key="Neuve-Sainte-Geneviève">rue Neuve-Sainte-Geneviève\</name>\</rs>, entre \<rs type="place-quartier" id="pl3">le\<name type="place-quartier" key="latin">quartier latin\</name>\</rs> et le \<rs type="place-rue" id="pl4">\<name type="place-rue" key="Saint-Marceau">faubourg Saint-Marceau\</name>\</rs>\</rs>.\</seg>\<seg>\<rs type="org-oeuvre" id="or2">Cette pension, connue sous le nom de la\<name type="org-oeuvre" key="Maison-Vauquer">Maison-Vauquer\</name>\</rs>, admet également des hommes et des femmes, des jeunes gens et des vieillards, sans que jamais la médisance ait attaqué les moeurs de \<rs type="org-oeuvre" id="or3">ce respectable établissement\</rs>.\</seg>...\</p> |

**Figure 2.** *« Goriot » (extrait) : TEI structure et sémantique (Bruneseaux et al. 1997)*

---

[7] http://visl.sdu.dk/visl/fr/parsing/automatic/

| Segmentation |
|---|
| `<word id="word_27">Madame</word>`<br>`<word id="word_28">Vauquer</word>`<br>`<word id="word_29">,</word>`<br>`<word id="word_30">née</word>`<br>`<word id="word_31">De</word>`<br>... |

**Figure 3.** *« Goriot » (extrait) : texte segmenté*

| Morpho-syntaxe 1 | | | | |
|---|---|---|---|---|
| 1 | Madame | madame | NCFIN | Ncf. |
| 2 | Vauquer | Vauquer | NPI | Np.. |
| 3 | , | , | PCTFAIB | Ypw |
| 4 | née | naître | VPARPFS | Vmpasf |
| 5 | De | de | PREP | Sp |
| 6 | Conflans | Conflans | NPI | Np.. |
| 7 | , | , | PCTFAIB | Ypw |
| 8 | est | être | VINDP3S | Vmip3s |
| 9 | une | un | DETIFS | Da-fs-i |
| ... | | | | |

**Figure 4.** *« Goriot » (extrait) : analyse morpho-syntaxique (Cordial)*

| Morpho-syntaxe 2 |
|---|
| `<w span="word_27"   msd="SBC:_:s"   lemma="madame"/>`<br>`<w span="word_28"   msd="SBP"   lemma="vauquer" />`<br>`<w span="word_29"   msd=" " lemma="," />`<br>`<w span="word_30"   msd="ADJ2PAR:f:s"   lemma="naître" />`<br>`<w span="word_31"   msd="PREP" lemma="de" />`<br>`<w span="word_32"   msd="SBP"   lemma="conflans" />`<br>`<w span="word_33"   msd=" " lemma="," />`<br>`<w span="word_34"   msd="ECJ:3p:s:pst:ind"   lemma="être:3g" />`<br>`<w span="word_35"   msd="DTN:m:s"   lemma="un" />`<br>... |

**Figure 5.** *« Goriot » (extrait) : analyse morpho-syntaxique (WinBrill, sortie XML)*

| Syntaxe | |
|---|---|
| S:prop("Madame_Vauquer") | Madame_Vauquer |
| , | |
| Vm:v-pcp2('naître' F S) | née |
| DN:pp | |
| =H:prp("de") | De |
| =DP:prop("Conflans") | Conflans |
| , | |
| Vm:v-fin('être' PR 3S IND) | est |
| Cs:np | |
| =DN:art('une' \<idf\> F S) | une |
| ... | |

**Figure 6.** *« Goriot » (extrait) : analyse syntaxique (VISL-FrAG, Bick 2003)*

| Syntaxe et sémantique |
|---|
| prop. principale |
|     * Sujet [est] |
|         Madame (polysémique : titre donné aux femmes nobles) |
|         Sous-groupe+ |
|             Vauquer |
|         , |
|         née (polysémique : venir au monde) |
|         * Complément d'objet indirect [née] |
|             de |
|             Conflans |
| ... |

**Figure 7.** *« Goriot » (extrait) : analyse syntaxique et sémantique (Cordial)*

| Coréférence |
|---|
| Madame Vauquer, née De Conflans , est une vieille femme qui, depuis quarante ans, tient à Paris \<coref id="1"\>une pension bourgeoise\</coref\> établie rue Neuve-Sainte-Geneviève , entre le quartier latin et le faubourg Saint-Marceau . \<coref id="2" type="ident" ref="1"\>Cette pension\</coref\>, connue sous le nom de la Maison-Vauquer , admet également des hommes et des femmes, des jeunes gens et des vieillards, sans que jamais la médisance ait attaqué les mœurs de ce respectable établissement. |

**Figure 8.** *« Goriot » (extrait) : annotation coréférentielle (Vieira et al., 2005)*

*2.2. Articulation avec les métadonnées*

Les métadonnées associées à des données linguistiques en documentent le contenu, le format, l'historique de sa constitution et de ses révisions ainsi que les conditions de distribution. L'objectif principal de cette documentation est de faciliter l'identification et l'accès à la ressource, que ce soit à travers une interface de requête associée à une base de corpus ou par un référencement sur un serveur de métadonnées, tel que le serveur OLAC[8]. Contrairement aux données, qui peuvent éventuellement être soumises à des restrictions d'accès, les métadonnées sont, par principe, systématiquement libres d'accès. Parmi les initiatives majeures visant à inventorier les besoins en métadonnées spécifiques aux corpus linguistiques, la définition de l'en-tête de la TEI a joué un rôle précurseur. D'autres projets ont relayé l'initiative, en particulier OLAC, IMDI[9] et INTERA, avec la proposition d'un ensemble de métadonnées adaptées à divers types de ressources linguistiques (corpus oraux, écrits, multimodaux et lexiques) à partir du Dublin Core[10].

À l'organisation architecturale de la *FReeBank* en corpus, niveaux de description linguistique et ressources s'ajoute donc une documentation sous forme de méta-données. Ces métadonnées ont été choisies parmi celles proposées dans le cadre des initiatives mentionnées précédemment, en fonction de leur pertinence vis-à-vis des composantes de l'architecture sous-jacente et reposant crucialement sur la définition que nous avons donnée à chacune des composantes dans la section 2.1.

La composante *corpus* est caractérisée par les métadonnées permettant de documenter la couverture linguistique. Il s'agira d'informations globales relatives au contenu linguistique. Parmi celles-ci, mentionnons la (ou les) langue(s), des statistiques lexicales élémentaires (nombre de mots, de tours de paroles), ainsi que la caractérisation effective de la source, qu'il s'agisse de données écrites (ouvrage d'origine) ou orale (caractérisation des conditions de recueil).

La composante *niveau de description* est caractérisée par des métadonnées permettant de qualifier les choix linguistiques ou éditoriaux correspondant à la contribution spécifique du niveau de description, sans dupliquer ce qui a été décrit au niveau du corpus. Il pourra s'agir des choix de transcription, d'un schéma d'annotation particulier, de l'identification de la personne ou de l'outil à l'origine de ce niveau de description, et, le cas échéant, du niveau de description sur lequel s'ancre éventuellement le niveau courant (cas d'une annotation externe). En lien avec la gestion de différentes versions d'un même niveau de description, certaines de ces métadonnées sont commentés de façon plus détaillée dans la section 3.2.1.

Enfin, la documentation de la composante *ressource* sera limitée à l'identification de l'entité responsable du dépôt de la ressource en question, des dates de dépôt, ainsi que des conditions de diffusion spécifiques associées.

---

[8] http://www.language-archives.org/
[9] http://www.mpi.nl/IMDI/
[10] http://dublincore.org/

*2.3. Modélisation d'une archive de corpus linguistiques*

Nous proposons maintenant d'intégrer les différents éléments présentés dans les sections précédentes au sein d'un modèle unique qui serve de base de représentation, mais aussi de déploiement, d'une archive ouverte de corpus linguistiques. Nous souhaitons ainsi, au-delà de l'expérience de la *FReeBank*, offrir un cadre conceptuel intégré de descriptions d'archives linguistiques au sens large. Le modèle, schématisé dans Figure 9, s'articule autour des trois composantes que sont le *corpus*, le *niveau de description* et la *ressource*. L'organisation de ces trois composantes entre elles fait apparaître de fait deux vues complémentaires attachées à la notion de corpus : une vue conceptuelle d'abord, qui permet de voir un corpus comme formé d'un certain nombre (*0 à n*) niveaux de descriptions, identifiant l'organisation des différentes représentations linguistiques (description et/ou analyse du matériau linguistique) ; une vue opérationnelle ensuite, qui décompose le corpus en ressources (*0 à n*), correspondant aux unités de dépôt et d'archivage.

Dans ce cadre, le corpus est le point d'entrée unique à l'intérieur de l'archive et peut, comme on le voit, n'être associé à aucune ressource ou niveau de description. On modélise ainsi le cas particulier d'un serveur de méta-données, qui ne fait que répertorier des ressources existantes, numérisées ou non (comme, par exemple, le portail OLAC). Du point de vue des niveaux de description, le modèle intègre la possible dépendance d'un niveau donné vis-à-vis d'un ou de plusieurs autres niveaux. Cette dépendance permet, comme nous l'avons vu, de gérer la logique d'organisation de certaines informations linguistiques entre elles, mais aussi de donner une base solide à la définition du caractère primaire ou secondaire d'un niveau de description donné.

Par ailleurs, le lien entre niveau de description et ressource explicite la contrainte qu'il ne puisse y avoir de ressource déposée qui ne contienne au moins un niveau de description. A l'inverse, nous n'imposons pas qu'un niveau de description soit explicitement associé à une ressource effective, considérant qu'une base de méta-données pourrait identifier des annotations disponibles pour un corpus donné, sans pour autant les archiver physiquement.

Enfin, la décomposition en niveaux de description d'un corpus donné est indépendante de l'intégration ou non de tout ou partie de ces niveaux de description en une seule et même représentation (paradigme de l'annotation interne). Quelle que soit la représentation adoptée, nous faisons l'hypothèse que les niveaux de description concernés sont clairement identifiés et documentés en tant que tel.

Combinées aux composantes du modèle d'architecture, les métadonnées définissent un modèle de documentation complet associé à la *FReeBank*, dans la ligné des travaux de modélisation menés au sein du TC 37/SC 4 (Ide et Romary, 2004a). D'un point de vue opérationnel, ces métadonnées se présentent sous forme d'en-têtes TEI, générés automatiquement à partir de l'interface de saisie lors du dépôt d'une ressource. Un certain nombre de métadonnées sont renseignées par le dépositaire, d'autres seront complétées par des informations générées automatiquement par la base (dates de dépôt d'une ressource, taille d'un corpus, statistiques sur les unités de segmentation élémentaires, etc.). Une version simplifiée de cette option est intégrée dans l'implémentation actuelle la *FReeBank*[11,12].

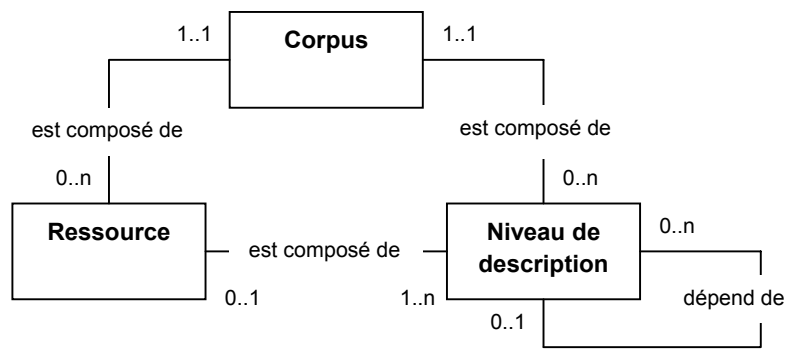

**Figure 9.** *Architecture pour une archive de corpus linguistiques*

## 3. Alimentation de la *FReeBank*

### 3.1. *Corpus et niveaux de description disponibles*

Le prototype de la *FReeBank* accueille d'ores et déjà certains corpus collectés et annotés dans le cadre du projet *Ananas* (Salmon-Alt, 2002). Lors de cette collecte, l'accent a été mis sur des corpus appartenant à des genres variés, et dans la mesure du possible, accessibles librement à des fins de recherche : textes littéraires, journalistiques, techniques, scientifiques et administratifs. Actuellement, nous disposons de données d'environ un million de mots. Les niveaux de description associés aux corpus sont présentés de façon plus détaillée dans la suite : segmentation en unités de référence, analyse structurelle en unités textuelles fondamentales (textes, titres, paragraphes, phrases, divisions), double voire triple annotation morpho-syntaxique, analyse syntaxique et annotation de certaines anaphores. Le Tableau 1 donne une vue synthétique des données actuellement disponibles dans la *FReeBank*. (cf. note 12).

---

[11] http://www.atilf.fr/freebank
[12] Note aux relecteurs : Il existe un prototype d'implémentation de la *FReeBank*, accessible sur http://www.loria.fr/projets/freebank. Ce prototype fonctionne pour les opérations élémentaires (dépôt et téléchargement de corpus), mais n'intègre ni totalité des concepts développés ici, ni la totalité des ressources disponibles et décrites dans la section 3.1. Nous sommes en train de redéployer le site à l'ATILF et y publierons les ressources avant l'été 2005. En attendant, l'ATILF met à disposition toutes les ressources décrites sur simple demande.

| *Titre du corpus Taille en mots Genre* | *Description* | *Constitution et contributions majeures (Institution, projet, personnes)* | *Niveaux de description disponibles* |
|---|---|---|---|
| **Père Goriot** 100.000 littéraire | extraits « Le Père Goriot » de H. Balzac | LORIA/LED (F. Bruneseaux) | TEI, segmentation, morpho-syntaxe, syntaxe |
| **Vittoria** 10.000 littéraire | « Vittoria Accoramboni, duchesse de Bracciano » (H. Beyle) | LIMSI (A. Popescu-Belis, I. Robba) | TEI, segmentation, morpho-syntaxe, syntaxe, anaphores |
| **Alice** 30.000 littéraire | extraits « Alice au pays des merveilles » (L Carroll) | LORIA/LED (Silfide) | TEI, segmentation, morpho-syntaxe |
| **Les Misérables** 100.000 littéraire | extraits « Les Misérables » (V. Hugo) | ATILF (Frantext) | TEI, segmentation, morpho-syntaxe |
| **Le Monde 1997** 65.000 journalistique | extraits « Le Monde » (09/1987) | ATILF (Parole) LORIA/LED (H. Manuélian) | TEI, segmentation, morpho-syntaxe, syntaxe, anaphores |
| **Le Monde Diplo** 120.000 journalistique | extraits « Le Monde Diplomatique » (1998) | LORIA/LED (Silfide) U de Grenoble 3 ( C. Clouzot) | TEI, segmentation, morpho-syntaxe, anaphores |
| **Est Républicain** 20.000 journalistique | extraits « L'Est Républicain » (2001) | L'Est Républicain ATILF (Ananas) | TEI, segmentation, morpho-syntaxe |
| **JOC** 60.000 administratif | « Journal officiel de la CE » (questions-réponses des parlementaires, 1992) | LORIA/LED ATILF (CommonRefs) | TEI, segmentation, morpho-syntaxe, syntaxe, anaphores |
| **Constitution** 7.000 administratif | Constitution de la France (version du 4/10/1958) | ATILF (Ananas) | TEI, segmentation, morpho-syntaxe |
| **Charte Dentistes** 7.000 administratif | Charte professionnelle des chirurgiens-dentistes | ATILF (Ananas) | TEI, segmentation, morpho-syntaxe |
| **MAIF** 7.000 administratif | constats d'accidents de voiture | transmis par GREYC (P. Enjalbert) | TEI, segmentation, morpho-syntaxe, syntaxe, anaphores |
| **Journal CNRS** 25.000 scientifique | extraits du « Journal du CNRS » (CNRS Editions) | ATILF (Ananas) | TEI, segmentation, morpho-syntaxe |

**Table 1.** *Corpus disponibles dans la FReeBank (avant l'été 2005, cf. note 12)*

*3.1.1. Segmentation en unités de référence*

L'objectif d'une première étape de traitement était de constituer les données correspondant au niveau de description primaire : il s'agit de données de référence, permettant de reconstituer la couverture des corpus et fournissant le point d'ancrage pour les autres niveaux de description. Les corpus actuels de la *FReeBank* fournissent à ce niveau un découpage en unités linguistiques minimales (Figure 3). Les composants des lexèmes complexes, des noms propres et des déterminants contractés ont été séparés. Ce choix aura permis d'ancrer de façon optimale les informations issues d'analyses ultérieures par une identification précise des unités linguistiques en question : une analyse morphologique qui décompose les déterminants (*au* en *à + le*) trouvera les deux points de référence nécessaires, alors qu'il n'y aura pas de perte d'informations dans le cas contraire. De même, au niveau syntaxique, ce choix permettra de distinguer avec précision le début d'un groupe prépositionnel de celui du groupe nominal imbriqué ([*à* [*le château*]]). Ces deux aspects se révèleront particulièrement importants pour l'annotation anaphorique. (section 3.1.5.)

*3.1.2. Niveau de description structurelle*

Nous ne souhaitions perdre aucune des informations présentes dans les corpus de départ, qu'il s'agisse d'informations linguistiques ou d'informations inhérentes à la structuration du texte, essentiellement en phrases, paragraphes, sections et titres (Figure 2). A partir d'un inventaire des différents éléments structuraux dans les corpus de la base, nous avons défini des schémas d'annotation suivant les recommandations de la TEI. Pour les ressources sans marquage explicite de la structure initiale, nous avons généré la description des principales unités structurelles repérables de façon automatique : sections, paragraphes, titres, phrases. La rétro-conversion des corpus qui étaient déjà annotés en paragraphes, tours de parole, titres, sections et/ou discours directs comporte en plus une phase de re-synchronisation des pointeurs avec les unités de référence (section 3.1.1.), travail qui est actuellement en cours.

*3.1.3. Niveau de description morpho-syntaxique*

Tous les corpus ont fait l'objet d'une annotation morpho-syntaxique. Ce niveau de description identifie des unités linguistiques pertinentes d'un point de vue morpho-syntaxiques (égaux ou supérieurs aux unités de référence décrits en 3.1.1.) et leur attribue une catégorie grammaticale, des informations flexionnelles et un lemme. Par défaut, ces informations sont celles issues d'une analyse effectuée par *Cordial*, dont le résultat a été converti en annotation XML externe (sur le même principe que celle de la Figure 5). Le lien avec les segments de référence se fait grâce à un pointeur sur un ou plusieurs de ces segments. Le cas d'une référence à plusieurs segments se présente lorsque plusieurs unités minimales ne forment qu'une seule entité morphologique, par exemple pour les déterminants contractés ou des mots composés.

En plus de l'analyse effectuée par *Cordial*, nous avons sauvegardé des annotations morpho-syntaxiques éventuellement préexistantes. Par ailleurs, un étiquetage supplémentaire – basé sur le *DecisionTreeTagger* (Schmid 1994) et des grammaires locales de désambiguïsation (Bick 2003) – est réalisé au cours de l'analyse syntaxique. De ce fait, le niveau de description morpho-syntaxique de la *FReeBank* est l'illustration par excellence de l'utilité de la mise oeuvre du principe de représentation externe. Sans cette solution, la maintenance de plusieurs versions d'annotation morpho-syntaxique pour un même corpus aurait automatiquement nécessité la maintenance de plusieurs niveaux de description primaires, avec, comme conséquence majeure, la perte de la garantie de la cohérence de la couverture linguistique, et donc, in fine, de l'identité du corpus.

*3.1.4. Niveau de description syntaxique*

L'annotation syntaxique était une étape essentielle et critique, puisqu'elle était destinée à permettre l'extraction automatique des groupes nominaux et pronominaux, intervenant ultérieurement dans l'annotation anaphorique. L'identification des constituants syntaxiques ainsi que de leurs fonctions s'est faite à l'aide de l'analyseur *VISL-FrAG* (Bick, 2003), accessible librement en ligne[13]. Il s'agit d'un système multi-niveaux hybride, combinant approche probabiliste, grammaire à base de contraintes et grammaire de structures phrastiques. L'entrée, sous forme de texte brut, est d'abord étiquetée par le *DecisionTreeTagger,* puis corrigée et désambiguïsée à l'aide d'un lexique morphologique et de règles locales contextuelles. La sortie est traitée par un système hiérarchique de grammaires de contraintes, ajoutant et désambiguïsant des étiquettes pour les formes et fonctions syntaxiques. Ensuite intervient une grammaire de structures phrastiques, basée non pas sur des terminaux traditionnels (mots), mais sur les fonctions syntaxiques. L'avantage d'un tel système hybride est de combiner la robustesse des grammaires par contraintes avec la profondeur des grammaires de structures phrastiques. Afin de réduire les ambiguïtés liées aux structures coordonnées et au rattachement des groupes nominaux, une grammaire spécialisée pour les attachements est utilisée à un niveau intermédiaire. La dernière étape, la sélection automatique de l'arbre d'analyse, est optionnelle et peut être replacée par un choix semi-manuel. Pour l'instant, l'annotation syntaxique a été réalisée pour un quart des corpus disponibles de la *FReeBank*. Un dixième de ces données (20.000 mots environs) a d'ailleurs fait l'objet d'une correction et d'une validation manuelle (section 3.2.1).

*3.1.5. Niveau de description coréférentielle et anaphorique*

L'annotation de la coréférence et des anaphores demande d'abord un marquage des expressions entrant potentiellement dans des relations coréférentielles ou anaphoriques. Cela concerne les expressions (pro)nominales, mais aussi d'autres types de constituants tels les groupes verbaux, les phrases ou même les paragraphes. En pratique, nous avons opté pour une extraction de tous les groupes (pro)nominaux de longueur maximale, puis un filtrage, excluant essentiellement les pronoms personnels autres que ceux de 3$^e$ personne, possessifs, explétifs, vides, réflexifs, ainsi que les groupes nominaux appositifs, les noms suivant directement une préposition (*de cyclisme, en juillet*) et les expressions nominales temporelles (*ce matin, le lendemain*). Pour les corpus n'ayant pas fait l'objet d'une analyse syntaxique préalable, l'ensemble de ces constituants a été sélectionné manuellement. Pour les autres corpus, ces informations ont été extraites automatiquement, puis filtrées semi-automatiquement.

---

[13] http://sandbox.visl.sdu.dk/visl/fr/

La deuxième étape a été l'annotation manuelle des liens coréférentiels et anaphoriques. Après plusieurs phases d'expérimentation sur différents corpus et avec différents annotateurs, nous avons obtenu le meilleur taux d'accord entre annotateurs en procédant par une annotation en plusieurs passes. D'abord, nous avons demandé aux annotateurs de séparer les anaphores coréférentes des autres anaphores. Ensuite, les anaphores coréférentes « fidèles » (à tête identique) ont été séparées des anaphores coréférentes « infidèles » (à tête divergente). Enfin, il s'agissait d'identifier, parmi les anaphores non coréférentes, celles ayant une relation bien typée avec l'antécédent (tout-partie, ensemble-membre, etc.).

Actuellement, les données de la *FReeBank* annotées en anaphores et/ou coréférence approchent les 180.000 mots (répartis sur neuf corpus). Elles comportent en tout 18.000 expressions référentielles et 8.500 liens référentiels. Environ la moitié de ces données provient de projets antérieurs (Bruneseaux 1997, Popescu-Belis et al. 1998, Clouzot et al. 2000) rétro-converties, l'autre moitié ayant été annotée par nos soins (Vieira et al., 2005).

### 3.2. *Questions de « bonne pratique » éditoriale*

La constitution des données actuelles de la *FReeBank* a été accompagnée par des décisions éditoriales concernant, d'une part, la validité linguistique des annotations, et d'autre part, la normalisation de leur représentation. Sans vouloir imposer nos choix à d'autres contributions venant potentiellement alimenter la *FReeBank*, nous considérons ces questions comme importantes lors de la mise en oeuvre d'une archive de ressources linguistiques réutilisables. C'est aussi dans cet esprit que nous avons conçu le prototype de la *FreeBank* comme une plate-forme de distribution de corpus à vocation générique et ouverte, et que le modèle formel d'architecture, développé dans la section 2., nous fournit le point de départ pour l'implémentation d'un portail évolutif, capable d'intégrer et de documenter des contributions de natures très diverses soumises en ligne, qu'il s'agisse de nouveaux corpus ou de mises à jour du matériau existant, d'annotations internes ou externes, ou de formats normalisées ou propriétaires.

*3.2.1. Des contraintes sur la validité linguistique à la gestion des versions*

En prolongement du débat sur le degré d'interprétation dans une annotation de corpus (cf. la discussion de la section 2.1.2.) se pose la question de la validité linguistique des données déposées à la *FReeBank*. Ayant conscience que la frontière entre variante interprétationnelle et analyse linguistique invalide peut être floue dans certains cas, nous proposons toutefois de faire reposer cette distinction sur l'accord entre jugements humains : si, pour une variante interprétationnelle, il peut y avoir divergence entre plusieurs jugements humains ou hésitation pour un même annotateur, une analyse linguistique fausse sera identifiée en tant que telle de façon convergente. Le premier cas est illustré par l'annotation anaphorique (Figure 10) : ici, un même annotateur humain hésite entre l'attribution de deux antécédents différents (*les technologies de l'information* vs. *les technologies de l'information* et *l'infosphère*) à une même anaphore pronominale (*elles*). La Figure 11 montre un cas de fausse analyse morphologique de la forme *suis*, analysée en tant que forme fléchie de *suivre* plutôt que de *être*.

```
Incertitude technique, d'abord, tant il est difficile de discerner les implications à moyen et long terme
de l'explosion <referentialMarkable id="m_1">des technologies de l'information </struct> et de
l'émergence d'<referentialMarkable id="m_2">une infosphère</struct> dont M. de Saint-Germain
souligne - incertitude supplémentaire - qu'<referentialMarkable id="m_3">elles</struct> sont pilotées
par le marché civil.
<alt>
    <referentialLink referentialSource="id(m_3)" referentialTarget="id(m_1),id(m_2)"/>
    <referentialLink referentialSource="id(m_3)" referentialTarget="id(m_1)"/>
</alt>
```

**Figure 10.** *Variante d'interprétation (« Le Monde Diplo », Clouzot et al., 2000)*

```
    <w  lemma="ce">C'</w>
    <w  lemma="être:3g">est</w>
    <w  lemma="lui">moi</w>
    <w  lemma="qui">qui</w>
    <w  lemma="suivre:3g">suis</w>
    <w  lemma="le">l'</w>
    <w  lemma="auteur">auteur</w>
    <w  lemma="de">de</w>
    <w  lemma="ton">ta</w>
    <w  lemma="joie">joie.</w>
```

**Figure 11.** *Analyse fausse (« Goriot », WinBrill)*

Le modèle d'architecture de la *FReeBank* étant générique, le degré de validité linguistique des données déposées dépend uniquement des décisions de nature éditoriales. Dans la perspective d'une plate-forme évolutive et ouverte à l'enrichissement progressif par des contributions de la communauté scientifique, nous avons fait le choix de ne pas imposer une validation linguistique des données *a priori*. Cela signifie que certaines informations résultant d'une analyse linguistique humaine ou automatique sont potentiellement invalides ou sujet à des variantes interprétationnelles.

Concernant les informations invalides, il peut s'agir de fausses analyses automatiques, par exemple de certaines sorties d'un analyseur syntaxique, ou d'erreurs commises lors d'une annotation manuelle. Nous posons toutefois le postulat que les données déposées présenteront toujours de l'intérêt : la majorité des informations restera valide et exploitable, aucune annotation de corpus (ni automatique, ni humaine) n'est infaillible, et de la sorte, la *FReeBank* donnera une image relativement fidèle de l'état de l'art dans l'annotation automatique et manuelle de corpus. Par ailleurs, nous espérons qu'une dynamique d'évolution et d'amélioration, concernant à la fois les outils TAL et les ressources, s'installera à partir des données déposées et donc accessibles à l'ensemble de la communauté : repérage et correction automatiques des analyses linguistiques insuffisantes ou invalides, travail sur des mesures d'évaluation pour l'accord inter-annotateur, fusion de résultats de plusieurs analyses, études linguistiques manuelles sur certains phénomènes. Ce type de travaux pourra donner lieu à une stratification d'un même niveau de description : une analyse syntaxique automatique à large couverture pourra par exemple faire l'objet d'une correction manuelle selon le paradigme « transverse », c'est-à-dire focalisée sur certains points bien identifiés. Wallis (2003) a en effet constaté qu'une telle approche de la correction manuelle est à la fois moins coûteuse et plus fiable qu'une correction qui se veut exhaustive. Selon notre propre expérience, la correction manuelle exhaustive de la sortie de l'analyseur syntaxique *VISL-FrAG* (Bick, 2003) pour un extrait de 20.000 mots a demandé un homme/mois. Or, cette correction, bien qu'effectuée par une linguiste entraînée, a introduit de nouvelles erreurs. Cela signifie qu'il faudrait prévoir, pour aboutir à des données supposées « valides », au moins une double correction, une phase de comparaison systématique, une phase de concertation, puis une phase de prise de décision par une tierce personne en cas de désaccord subsistant. L'extrapolation de notre expérience à un corpus de 1 million de mots aboutirait alors à un coût approximatif de 150 hommes/mois, coût qui est par ailleurs réaliste comparé à celui du projet SALSA (annotation de rôles sémantiques), procédant précisément selon ces principes (Erk et Pado, 2004). En considérant que l'état de l'art est en évolution très rapide, il nous paraîtrait vain de mettre en œuvre de tels moyens dans l'espoir de créer un possible corpus de référence qui serait, à peine produit, immédiatement dépassé.

Une partie des données linguistiques de la *FReeBank*, même linguistiquement valide, sera toujours sujette à des variantes interprétationnelles. Bien que Leech (1993) préconise l'usage de schémas d'annotation aussi « neutres » que possibles par rapport à des théories particulières, on ne peut exclure des variantes dues à des approches théoriques divergentes. D'autres variantes proviennent plus simplement de différentes analyses automatiques (cf. le niveau d'annotation morpho-syntaxique des corpus noyau de la *FReeBank*) ou de la possibilité de plusieurs interprétations humaines concurrentes (« ambiguïtés »). Il s'agit là d'un phénomène tout à fait régulier pour les langues naturelles, qu'il n'y aurait aucun intérêt à exclure de la *FReeBank*. Au contraire, leur marquage explicite peut s'avérer extrêmement utile, par exemple pour la définition de clés lors de campagnes d'évaluation d'outils TAL ou pour la validation d'approches cognitives.

Qu'il s'agisse de données visant à « corriger » des données existantes ou de données sujettes à des interprétations multiples, la gestion de ce matériau dans la *FReeBank* passe obligatoirement par une documentation fine sous forme de méta-données appropriées (cf. section 2.2). Plus précisément, il s'agit de spécifier des liens entre données parallèles relevant à la fois d'un même corpus et d'un même niveau de description. Les principaux cas de figure relevés précédemment – correction exhaustive, correction transverse, co-existence de plusieurs analyses humaines ou automatiques – peuvent en effet se caractériser à partir d'une combinaison de métadonnées subordonnées au niveau de description. Parmi ceux-ci, deux revêtent d'une importance particulière : la granularité du niveau de description et la validation linguistique explicite.

La granularité d'un niveau de description caractérise la nature et l'étendue des phénomènes linguistiques pris en compte et s'exprime par exemple par référence à une DTD. Toutefois, elle devra être formalisée à terme sous forme d'une référence à une sélection de catégories de données du registre des catégories de données[14] (Ide et Romary, 2004a). Par rapport à des données préexistantes dans la *FReeBank*, une nouvelle soumission relative à ces données (même corpus, même niveau de description) peut se caractériser par exemple par une granularité égale, plus fine ou différente. À granularité égale, il s'agira du dépôt d'une *version parallèle* : c'est par exemple le cas d'un nouvel étiquetage morpho-syntaxique automatique ou d'une deuxième annotation manuelle des pronoms personnels anaphoriques. À granularité plus fine, les données soumises fourniront une *version parallèle enrichie* : par exemple une annotation morpho-syntaxique sous-catégorisant plus finement les adverbes ou une annotation anaphorique étendue aux descriptions définies. À granularité différente, il s'agit tout simplement de *versions supplémentaires* (non parallèles).

La validation, elle, consiste en l'affirmation explicite, de la part d'un annotateur, d'un dépositaire ou d'un administrateur de la *FReeBank*, de la validité linguistique de données soumises. Lorsque la couverture et le niveau de description des données soumises correspondent à une version préexistante de la *FReeBank*, une *version validée* peut constituer de fait une correction exhaustive ou transverse. Il s'agira d'une correction exhaustive, lorsque la granularité de la version validée est égale ou supérieure à la version précédente. A granularité inférieure ou différente, il s'agira d'une correction transverse.

---

[14] http://syntax.loria.fr

Les métadonnées liées à la granularité des niveaux de description, combinées à la notion de validation linguistique explicite suffisent donc à caractériser tous les cas de co-existence de plusieurs analyses humaines ou automatiques ainsi que les cas des corrections exhaustives ou partielles relatives à des données initialement non validées. Nous pensons qu'offrir la possibilité de gérer convenablement ces variations est un élément essentiel pour qu'une archive telle que la *FReeBank* reflète au mieux les aspirations d'une large communauté scientifique.

*3.2.2. Représentation normalisée des données linguistiques*

Pour la représentation des données noyau de la *FReeBank*, nous nous sommes fixés comme priorité de garantir la compatibilité avec les standards internationaux, en mettant en oeuvre les recommandations pour la représentation de ressources linguistiques actuellement en cours de développement au sein du TC 37/SC 4 de l'ISO et relayées en France par l'initiative RNIL[15]. Celles-ci concernent plus particulièrement les niveaux de description correspondant aux « annotations » linguistiques au sens de McEnery et Wilson (1996), introduites dans la section 2.1. Dans cette optique, le processus d'annotation linguistique consiste, schématiquement, en l'identification d'unités pertinentes d'un point de vue linguistique, puis éventuellement, en la caractérisation de ces unités par des traits linguistiques et/ou en l'établissement de liens entre ces unités. Le résultat d'une annotation linguistique se présente dans un format particulier (parenthésage, base de données relationnelle, balisage etc.). Or, certaines initiatives précédentes, par exemple issues des campagnes d'évaluation américaines (Gerber et al., 2002), ayant tenté de proposer des standards sous forme de jeux de balises, se sont heurtées à un inconvénient majeur qui fut le manque de flexibilité, puisque l'on imposait aux utilisateurs à la fois le modèle de données sous-jacent (définition des propriétés des unités et des liens) et le format de représentation (SGML/XML+DTD).

Dans la lignée d'autres initiatives, plus génériques (Mengel et all., 2000; Bird et Liberman, 2001), le *Linguistic Annotation Framework* (LAF), tel que défini par Ide et Romary (2004b) préconise une séparation claire entre le modèle des données et les formats de représentation, partant du principe que la standardisation de l'annotation linguistique doit s'effectuer au niveau conceptuel plutôt qu'au niveau représentationnel. LAF propose donc une modélisation conceptuelle des objets d'annotation sous forme d'un méta-modèle. Celui-ci reflète les propriétés structurelles des données d'annotation : il s'agit d'un graphe orienté représentant les d'unités pertinentes d'un point de vue linguistique ainsi que les contraintes régissant leur agencement. Les propriétés de ces unités sont caractérisées par des descripteurs linguistiques ou « catégories de données », dont la gestion est externalisée dans un registre de catégories de données en ligne (Ide et Romary, 2004a; cf. note 14). L'association d'un ensemble de catégories de données aux nœuds d'un méta-modèle donne lieu à un modèle de données pleinement spécifié. Si LAF ne se préoccupe pas prioritairement des formats d'instanciation (i.e. des schémas d'annotation concrets), il propose néanmoins un format de représentation pivot parfaitement isomorphe au modèle. Sera considérée comme étant conforme à LAF toute annotation pour laquelle il existe une procédure d'appariement avec ce format générique.

---

[15] http://pauillac.inria.fr/atoll/RNIL/home-fr.html

L'élaboration de modèles de données est actuellement en cours pour différents niveaux de description[16] : morpho-syntaxe (Clément et de la Clergerie, 2004) syntaxe (Brants et al., 2002, Ide et Romary, 2003), rôles sémantiques (Erk et Pado, 2004), coréférence et anaphores (Salmon-Alt et Romary, 2005). Pour chaque niveau, il s'agit d'identifier le méta-modèle et l'ensemble de catégories de données spécifiques à la description linguistique de ce niveau. Pour les données de la *FReeBank*, nous avons suivi ces initiatives, en essayant de les mettre en oeuvre de façon aussi systématique que possible.

Le TC 37/SC 4 travaille dès à présent à la définition d'un modèle générique dédié à l'annotation morpho-syntaxique (future norme ISO 24611 ; Clément et de la Clergerie, 2004). Ce modèle combine d'une part deux niveaux de segmentation et de catégorisation linguistique, et d'autre part un ensemble de catégories de données linguistiques permettant de qualifier les différents éléments du modèle. Une étude préliminaire a en particulier permis de regrouper une base de catégories morpho-syntaxiques intégrant la plupart des jeux d'étiquettes connus pour le français. L'expérience d'appariement de ces principes avec les données de la *FreeBank* a été particulièrement intéressante, puisqu'elle a permis de mettre à l'épreuve les recommandations sur les sorties de quatre formats morpho-syntaxiques différents. Les propriétés structurelles ainsi que les descripteurs du format retenu pour l'encodage homogène des informations morpho-syntaxiques de la *FReeBank* sont compatibles avec ces principes et permettront aux données d'être converties vers la future norme sans perte d'informations.

Concernant le niveau de description syntaxique, la sortie de l'analyseur était initialement conditionnée par les fondements théoriques du projet *VISL*[17]. Or, s'agissant d'un environnement multilingue pour l'analyse et l'apprentissage syntaxiques, les structures et descripteurs syntaxiques sous-jacents (« méta-modèle » et « catégories de données ») se sont révélés suffisamment génériques pour avoir été mis à l'épreuve sur d'autres langues (portugais, danois). Par ailleurs, le format de sortie de l'analyse a pu être converti de façon complètement automatique vers le format *TIGER*. Il s'agit là de l'un des formats convergents au niveau international pour l'encodage des formes et fonctions syntaxiques (Brants et al., 2002) que nous avons retenu pour la *FReeBank*, pour plusieurs raisons : il est conçu pour représenter les résultats d'une analyse syntaxique profonde (dépendances et constituants), il permet la mise en oeuvre d'une annotation externe

---

[16] L'annotation structurelle étant couverte par la TEI (Sperberg-McQueen et Burnard, 2002).
[17] http://beta.visl.sdu.dk

et il est accompagné par des outils de recherche de corpus arborés performants et libres[18].

Enfin, la prolifération des schémas pour l'annotation de la coréférence et des anaphores a donné lieu, depuis quelques années, à des tentatives de synthèse (Poesio, 2000 ; Salmon-Alt, 2001). L'idée-clé de ces initiatives est de proposer un modèle de données qui fédère les propriétés des schémas précédents et qui puisse être instancié selon les besoins concrets du codeur. Les grandes lignes de ces propositions – introduction d'éléments autonomes pour les expressions à annoter ainsi que dles liens entre celles-ci – ont été intégrées dans un travail visant la normalisation du codage de ce niveau de description (Salmon-Alt et Romary, 2005). La *FReeBank* sert actuellement de banc d'essai pour l'application des ces propositions à large échelle : toutes les annotations anaphoriques et coréférentielles existantes ont été converties automatiquement vers le modèle en cours de normalisation[19] et les catégories de données pertinentes pour la description des expressions ainsi que des liens ont été soumises au registre des catégories de données.

**4. La *FReeBank* : une vitrine des bonnes pratique et un espace de réflexion ?**

L'objectif clairement affiché des travaux que nous avons menés au cours des derniers mois autour du projet *FReeBank*, tant d'un point de vue théorique que de la réalisation effective d'une première plate-forme, est d'arriver à stabiliser un certain nombre de concepts fondamentaux reflétant l'état de l'art dans les domaines de la représentation et de l'archivage de corpus linguistiques. Cette vision intégrée de la création, de la gestion et de la diffusion de telles archives doit maintenant être relayée au sein de notre communauté, par des actions plus spécifiques dans différentes directions :

– contribution à l'édification et à la diffusion des normes et bonnes pratiques de représentation de données linguistiques : qu'il s'agisse de la TEI (dont la France est maintenant l'un des quatre sites hôte) ou de l'ISO (avec le rôle moteur du réseau RNIL), nous devons déterminer des actions (comparaison des pratiques existantes, tutoriaux, etc.) qui nous permettent de faire vivre les normes existantes et de les faire évoluer ;

– production d'échantillons reflétant ces bonnes pratiques : un travail concerté de différentes équipes de recherche devrait fournir des jeux de données annotés qui, dans la continuité des projets Ananas ou Asila, puissent servir d'exemple pour de nouveaux projets similaires ;

– identification d'un vrai champ de recherche autour des ressources linguistiques : il apparaît que la réponse aux problèmes de la gestion d'annotations multi-niveau ne peut se contenter d'une approche strictement technologique. Les futurs travaux en la matière doivent s'appuyer sur des approches conceptuelles solides, qui dépassent la simple connaissance d'XML ;

---

[18] http://www.ims.uni-stuttgart.de/projekte/TIGER/TIGERSearch/
[19] Les outils de conversion sont en libre accès sur http://www.atilf.fr/ananas.

− édification de principes généraux concernant l'archivage ouvert de ressources linguistiques : en espérant que d'autres projets d'archives linguistiques ouvertes voient le jour, il faut définir dès à présent des principes techniques minimaux en garantissant l'interopérabilité (interrogation multi-archive), mais surtout édifier une charte de telles archives qui puisse synthétiser l'opinion de notre communauté en matière de libre accès.

Les propositions faites ici ne sont qu'une étape pour avancer dans ces différentes directions et de fait, nous entrevoyons dès à présent de refondre complètement notre prototype initial (cf. note 12) dans une plate-forme qui, d'une part, intègre mieux le point de vue des utilisateurs dans l'organisation des mécanismes de dépôt et d'accès, et, d'autre part, s'articule plus étroitement avec les modèles de données associés aux différents niveaux de description (tant pour l'écrit que pour l'oral).

**Bibliographie**